\documentclass[11pt]{article}
\usepackage[final]{acl}

\usepackage{graphicx} 
\usepackage{xcolor}
\usepackage{colortbl}
\usepackage[most]{tcolorbox}

\usepackage{booktabs}
\usepackage{multirow}
\usepackage{adjustbox}
\usepackage{makecell}
\usepackage{tabularx}
\usepackage{longtable}
\usepackage{array}
\usepackage{subcaption}
\usepackage{wrapfig} 
\usepackage{wrapfig}

\usepackage[patch=none]{microtype}
\usepackage{caption}
\usepackage{stfloats}
\usepackage{lineno}
\usepackage{textcomp}
\usepackage[T1]{fontenc} 
\usepackage{balance}
\usepackage{enumitem} 
\usepackage{times}  
\usepackage[ruled,linesnumbered]{algorithm2e}
\usepackage{algpseudocode}
\usepackage{listings}
\usepackage{amsmath,amsfonts,bm}
\usepackage{amsthm}
\usepackage{verbatim}

\usepackage{pifont}
\usepackage{amssymb} 

\SetKwInput{KwInput}{Input}

\SetKwInput{KwOutput}{Output}

\SetKwComment{Block}{$\triangleright$\ }{}

\SetKwComment{Comment}{//  }{}

\DontPrintSemicolon

\SetKwIF{If}{ElseIf}{Else}{if}{}{else if}{else}{end if}%

\def\HiLi{\leavevmode\rlap{\hbox to \hsize{\color{gray!20}\leaders\hrule height .8\baselineskip depth .5ex\hfill}}}

\definecolor{darkblue}{rgb}{0, 0, 0.5}

\hypersetup{
  colorlinks=true,   
  citecolor=darkblue, 
  linkcolor=darkblue, 
  urlcolor=darkblue   
}

\theoremstyle{plain}

\theoremstyle{definition}

\theoremstyle{remark}

\def\eqref#1{equation~\ref{#1}}

\def\1{\bm{1}}

\DeclareMathAlphabet{\mathsfit}{\encodingdefault}{\sfdefault}{m}{sl}
\SetMathAlphabet{\mathsfit}{bold}{\encodingdefault}{\sfdefault}{bx}{n}

\title{\emph{rSIM}: Incentivizing Reasoning Capabilities of LLMs via Reinforced Strategy Injection}

\author{
  Sijia Chen$^{1\dagger}$ \quad \quad Baochun Li$^{2}$ \quad \quad  Di Niu$^{3}$ \\
  $^{1}$ The Hong Kong University of Science and Technology (Guangzhou) \\
  $^{2}$ University of Toronto, $^{3}$ University of Alberta
  \\
  {\tt\small sijiachen@hkust-gz.edu.cn, bli@ece.toronto.edu, dniu@ualberta.ca}
}

\begin{document}
\maketitle

\begin{abstract}


Large language models (LLMs) are post-trained through reinforcement learning (RL) to evolve into Reasoning Language Models (RLMs), where the hallmark of this advanced reasoning is ``aha'' moments when they start to perform \textit{strategies}, such as self-reflection and deep thinking, within chain of thoughts (CoTs). Motivated by this, this paper proposes a novel reinforced strategy injection mechanism (\emph{rSIM}), that enables any LLM to become an RLM by employing a small planner to guide the LLM's CoT through the adaptive injection of reasoning strategies. To achieve this, the planner (leader agent) is jointly trained with an LLM (follower agent) using multi-agent RL (MARL), based on a leader-follower framework and straightforward rule-based rewards. Experimental results show that \emph{rSIM} enables Qwen2.5-0.5B to become an RLM and significantly outperform Qwen2.5-14B. Moreover, the planner is generalizable: it only needs to be trained once and can be applied as a plug-in to substantially improve the reasoning capabilities of existing LLMs. In addition, the planner supports continual learning across various tasks, allowing its planning abilities to gradually improve and generalize to a wider range of problems.

\end{abstract}

\section{Introduction}
\label{sec:intro}

Large language models (LLMs) have been enhanced with advanced reasoning capabilities, evolving into Reasoning Language Models (RLMs) \cite{rlm-arxiv25} that solve problems through step-by-step reasoning, commonly referred to as chain-of-thought (CoT) \cite{cot-neurips22}. A key advancement in RLMs is their ability to integrate reasoning \textit{strategies}, such as self-reflection, decomposition, and deliberative thinking, into the CoT process, contributing to improved problem-solving accuracy.

Existing literature \cite{reft-acl24, llmrl-icml24}, especially the recent Group Relative Policy Optimization (GRPO) \cite{DeepseekR1-arxiv25}, primarily post-trains LLMs to evolve into RLMs purely through reinforcement learning (RL) algorithms. A hallmark of this evolution is the emergence of the ``aha moment'' where LLMs start to perform \textit{strategies} such as self-reflection within CoTs. We empirically found that the better performance of evolved RLMs correlates with the higher appearance frequency and more appropriate positions of these strategies compared to those in LLMs. However, our findings also indicate that when LLMs, particularly smaller ones, inherently lack the capacity to perform basic reasoning strategies, RL-based post-training is unable to transform them into capable RLMs.




Therefore, this paper proposes a reinforced strategy injection mechanism (\emph{rSIM}), which enables any LLM, in particular those as small as 0.5B, to evolve into an RLM with minimal or even no additional training. To achieve this, \emph{rSIM} introduces only an auxiliary planner that, at each reasoning step of an LLM, adaptively selects an appropriate strategy from a predefined set such as \textit{self-reflection}, \textit{decomposition}, and others, and injects it into the chain of thought (CoT) to guide the next step generation. Specifically, the \emph{rSIM} offers the following four key contributions:


\setlist[itemize]{left=0pt}
\begin{itemize}
    \item Injecting reasoning strategies adaptively into the CoT process of any LLM, including small and large ones, via a planner enables the LLM to directly gain the advanced reasoning ability like RLMs.
    \item Training the planner and LLM jointly as two agents under multi-agent RL (MARL) with the leader-follower algorithm \cite{leaderfollower-icml23} reinforces the planner's ability to inject strategies.
    \item Planners are pluggable, meaning that a planner trained on one task can be directly integrated with another LLM to enhance its reasoning ability on similar tasks.
    \item Planners support continual learning, as a planner in our \emph{rSIM} can be continuously trained across tasks to enhance its planning ability over a broader range of problems.
\end{itemize}


Our results across seven datasets covering mathematics, multi-task reasoning, and code generation verify the benefits of \emph{rSIM}. First, even small LLMs such as Qwen2.5-0.5B, when jointly trained with a planner (Qwen2.5-0.5B), can evolve into an RLM, achieving accuracy on par with Qwen2.5-14B on \texttt{MATH} \cite{MATH-arxiv21}. Second, using the trained planner as a plugin enables another LLM to outperform larger models by a significant margin without any additional training. Third, a planner trained on mathematics can be continuously fine-tuned on coding tasks such as \texttt{CodeAlpaca-20k} \cite{CodeAlpaca20K} to further guide an LLM like Qwen2.5-0.5B, achieving 17\% higher accuracy on code generation.

\section{Related Work}
\label{sec:related}

\textbf{Data distillation}. Knowledge distilled from large language models (LLMs) can be transferred to smaller models to enhance their performance \cite{kdsurvey-arxiv24}. In particular, \cite{DeepseekR1-arxiv25} verifies that larger LLMs can transfer their step-by-step reasoning abilities to smaller models by distilling chain-of-thought (CoT) \cite{cot-neurips22} samples, where LLM-generated reasoning traces serve as additional fine-tuning data. Fine-tuning with teacher-generated CoT outputs \cite{teachsmall-acl22, distillsystems-neurips24, dcots-emnlp24}, rationalizations \cite{SCoTD-acl23}, specialized reasoning skills \cite{symbolic-aaai25}, CoT and Program of Thought (PoT) \cite{mixdist-emnlp24}, or even incorrect CoT samples \cite{selfimprove-neurips22, vstar-colm24} can significantly improve the reasoning abilities of smaller models. In this paper, using the planner from our DPR framework as a plugin to improve the reasoning of smaller models can be viewed as transferring a human-level planning chain to them.

\textbf{Reinforcement learning}. Reinforcement learning (RL) \cite{RL-book} has been widely applied to decision-making tasks, as demonstrated by AlphaGo \cite{alphago-nature16} and AlphaZero \cite{alphazero-nature17}. RLHF \cite{RLHF-neurips22} first leveraged PPO \cite{PPO-17} to align models with human preferences. ReFT \cite{reft-acl24} pioneered the use of RL as a fine-tuning paradigm to enhance LLM reasoning performance. Building on this progress, DeepSeek-R1-Zero \cite{DeepseekR1-arxiv25} made a breakthrough by demonstrating that self-verification, reflection, and the ability to generate long CoTs in LLMs can be incentivized purely through RL, specifically using GRPO \cite{GRPO-arxiv24}. This enables smaller base LLMs, such as 3B and 7B models, to be trained directly to achieve reasoning performance comparable to stronger models. However, as noted by \cite{llmrl-icml24}, models are inherently constrained in their ability to explore CoT solutions beyond their existing capabilities. Consequently, weaker base models as small as 0.5B fail to benefit from RL training and lag behind in reasoning performance.

\textbf{Multi-agent LLMs}. Building on the development of using a single LLM as a planning or decision-making agent, multi-agent LLMs \cite{agentsall-tmlr24} based on MARL \cite{MADDPG-neurips17}, where multiple language models collaborate, have achieved significant progress in complex problem-solving \cite{llm-multiagent-survey-ijcai24, metagent-arxiv23, chain-agent-neurips24}. As highlighted by \cite{agentsall-tmlr24}, this structure has been successfully applied to practical tasks \cite{metagpt-arxiv23, agentsoftware-arxiv23, Roco-icra24}. To enhance performance, SOCRATIC \cite{SOCRATIC-acl23} trains a combination of two small distilled models to perform CoT reasoning in LLMs. In contrast to our DPR framework, SOCRATIC still relies on distilling the abilities of large models into smaller ones. Meanwhile, ReAct \cite{react-iclr23} enables LLMs to generate both reasoning traces and task-specific actions in an interleaved manner. Additionally, CORY \cite{cory-neurips24} fine-tunes LLMs as two autonomous agents, a pioneer and an observer, which leads to superior performance compared to standard PPO. However, our work is the first to decouple planning from the reasoning process and to build a two-agent system under the MARL framework that enables any LLM to benefit from the trained planner agent.

\section{Preliminary and Motivation}
\label{sec:preli}

\subsection{Reasoning Language Models via Reinforcement Learning}

Given a question \( q \), the reasoning language model (RLM), parameterized by \( \bm{\theta} \), generates a sequence of thoughts, denoted as \( \bm{o} = \left[\bm{z}_1, \bm{z}_2, \dots, \bm{z}_n\right] \), where each \( \bm{z}_i \) with \( i \in [1,\dots,n] \) is a textual description of the thought at the \( i \)-th reasoning step. The predicted solution \( \widetilde{y} \) is extracted from \( \bm{z}_n \) and compared with the ground truth \( y \).  In the reinforcement learning (RL) framework, represented as \( \left[\mathcal{S}, \mathcal{A}, \pi_{\bm{\theta}}, \mathcal{R}\right] \), the state \( \bm{s} \in \mathcal{S} \) corresponds to the tokens generated so far, while the action \( a \in \mathcal{A} \) is the next token sampled from the policy \( a \sim \pi_{\bm{\theta}}\left(a | s\right) \). The corresponding reward is denoted as \( r \in \mathcal{R} \).   To optimize the policy model \( \bm{\theta} \), the RL algorithm aims to maximize the expected cumulative reward, which is formulated as follows:
\begin{equation}
    \label{eq:rl}
    \nonumber
    J(\bm{\theta}) = \mathbb{E}_{\bm{s}, a \in \bm{o} \sim \pi_{\bm{\theta}_{old}}} \left[ \frac{\pi_{\bm{\theta}}(a | \bm{s})}{\pi_{\bm{\theta}_{old}}(a | \bm{s})} A^{\pi_{\bm{\theta}_{old}}} (\bm{s}, a) \right]
\end{equation}
where \( \bm{\theta}_{old} \) represents the most recent policy model. The advantage function \( A(\cdot) \) estimates how much better an action is compared to the expected return. To balance bias and variance, we employ Generalized Advantage Estimation (GAE), formulated as \( A^{\pi_{\bm{\theta}_{old}}}\left(s_t, a_t\right) = \sum_{l=0}^{T-t} (\gamma\lambda)^l \delta_{t+l} \) where \( \delta_t = r_t + \gamma V^{\pi_{\bm{\theta}_{old}}} (s_{t+1}) - V^{\pi_{\bm{\theta}_{old}}} (s_t) \), where \( t \) denotes the token index, and \( \gamma \in [0, 1] \) is the discount factor. In general, to mitigate the discrepancy between \( \bm{\theta} \) and \( \bm{\theta}_{old} \), methods such as Proximal Policy Optimization (PPO) \cite{PPO-17} and Group Relative Policy Optimization (GRPO) \cite{GRPO-arxiv24} incorporate a KL penalty term.

As demonstrated by GRPO, the policy model \( \bm{\theta} \) is trained using rule-based rewards, where the rewards encode human-defined rules to guide the model in adhering to these rules while improving reasoning accuracy.

\subsection{LLMs Without Inherent Reasoning Strategies Show Limited Improvement}
\label{subsec:motivation}

\begin{figure}[t]
    \centering
    \begin{subfigure}[b]{0.48\columnwidth}
        \includegraphics[width=\linewidth]{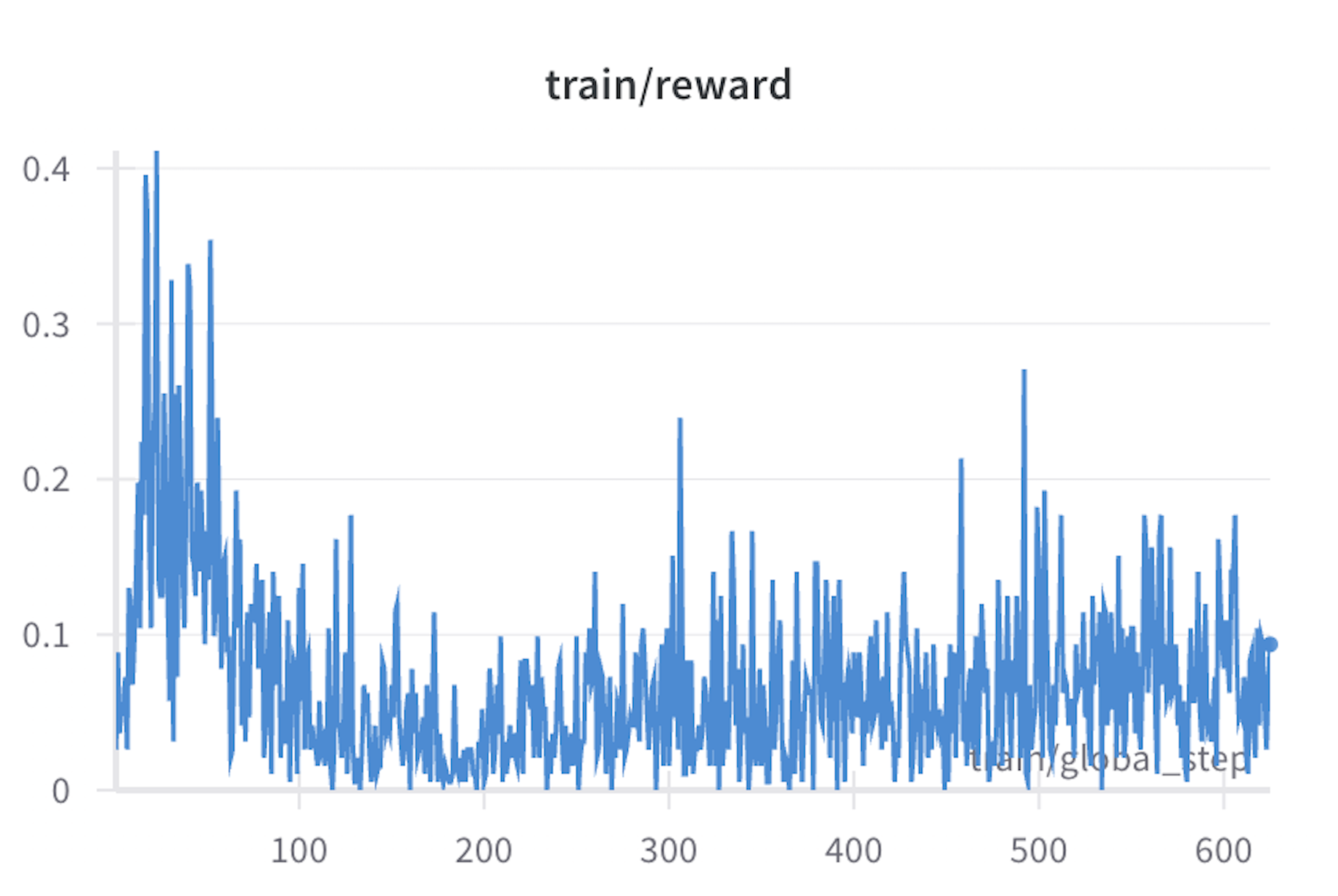}
        \caption{GRPO-based training failure on Qwen2.5-0.5B}
        \label{fig:r1}
    \end{subfigure}
    \hfill
    \begin{subfigure}[b]{0.48\columnwidth}
        \includegraphics[width=\linewidth]{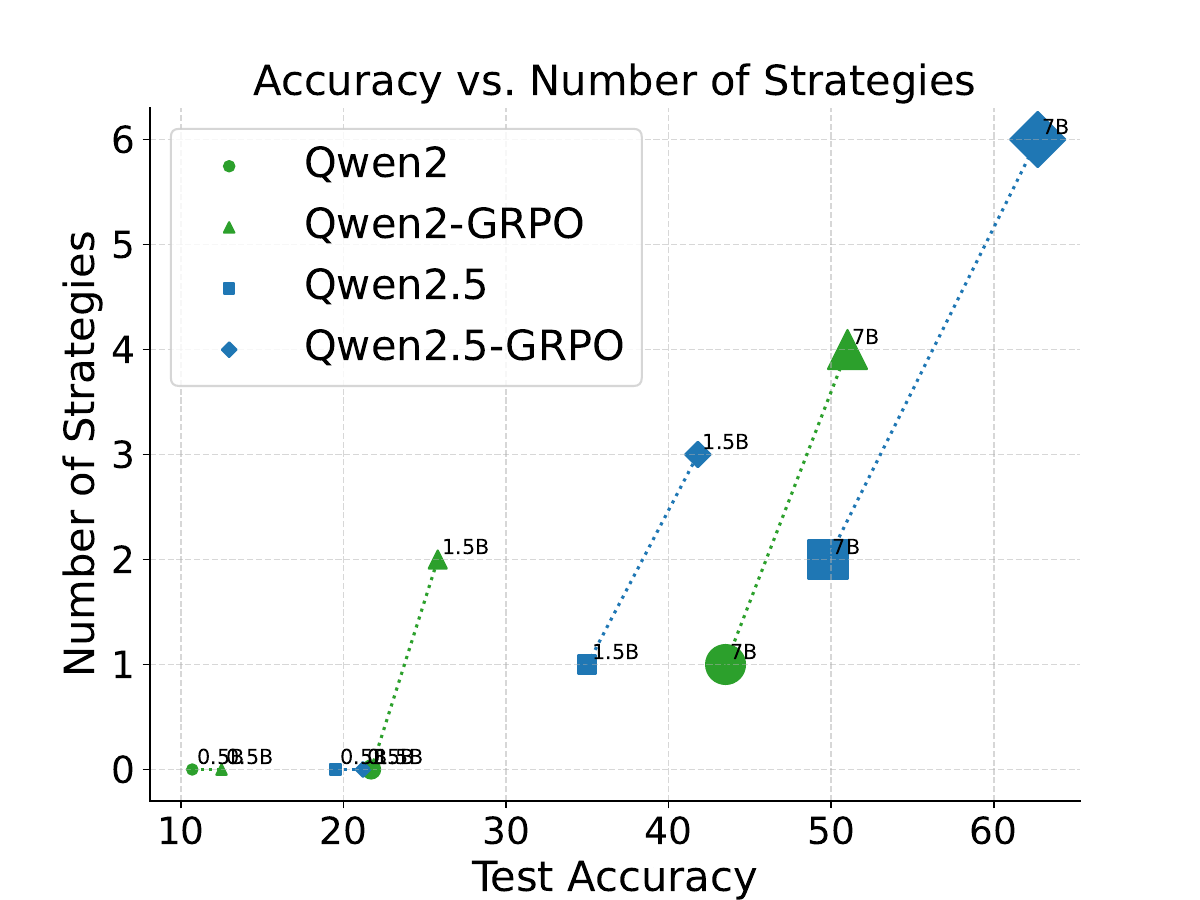}
        \caption{Accuracy vs. number of strategies across LLMs}
        \label{fig:r2}
    \end{subfigure}

    \caption{(a) Shows the absence of the ``aha'' moment in Qwen2.5-0.5B on the \texttt{MATH} dataset \cite{MATH-arxiv21}.
             (b) Compares performance of LLMs before and after training with GRPO. By detecting key words, we count the average number of reasoning strategies, as presented in Figure~\ref{fig:mainstructure}, that are used in answering each question.}
    \label{fig:motivation}
    \vspace{-4mm}
\end{figure}

Using a pure RL algorithm like GRPO, a base model is optimized to achieve an ``aha'' moment --- learning to exhibit more reasoning strategies, such as self-reflection and deep thinking, during training, leading to a sudden increase of model response length and ultimately to high accuracy. We argue that (1) when the base LLM is capable of performing reasoning strategies, reinforcement learning (RL) algorithms can optimize it to apply more strategies effectively during reasoning, and (2) there exists a positive correlation between the number of strategies and the accuracy. 

To prove this, we first define a set of the most commonly used \textit{strategies}: \textit{self-reflection, decomposition, deliberative thinking, validation, summarization, prioritization, continuation, sub-planning, and termination}. Following subsection \ref{subsec:counting}, we then count the number of times these strategies appear in the CoTs generated by the base LLMs and post-trained models.

As supporting evidence, we train LLMs of sizes 0.5B, 1.5B, and 7B from both Qwen2 and Qwen2.5 on the \texttt{MATH} dataset using the GRPO. First, as shown in Figure~\ref{fig:motivation}, when Qwen2.5-0.5B is used as the base model, the total reward initially increases to around 0.3 but then abruptly drops to 0 under GRPO training. Second, from Figure~\ref{fig:r2}, we observe that base models such as Qwen2-0.5B and Qwen2.5-0.5B, which lack inherent reasoning strategies (i.e., strategy count is 0), cannot be trained with RL to gain reasoning intelligence and show only limited improvement in accuracy. In contrast, models such as 1.5B and 7B, which demonstrate inherent reasoning strategies (with strategy counts greater than 0), can be further optimized through RL to apply more strategies during reasoning. More importantly, we observe that as the number of strategies increases, model accuracy also improves.

\section{Methodology}
\label{sec:framework}

Motivated by our observations in Subsection~\ref{subsec:motivation}, we introduce reinforced strategy injection mechanism (\emph{rSIM}), which allows a planner, implemented as a small-sized LLM, to guide another LLM by adaptively providing strategy instruction during CoTs. Through rejection, the planner can incorporate rich human-crafted knowledge and prior information, presented as those \textit{strategies}, to help the LLM effectively and directly gain advanced reasoning abilities and evolve toward becoming an RLM.

\subsection{Training Objective of the Two-Agent Framework}
\label{subsec:multiagent}

\begin{figure*}[t]
    \centering
    \includegraphics[width=\textwidth]{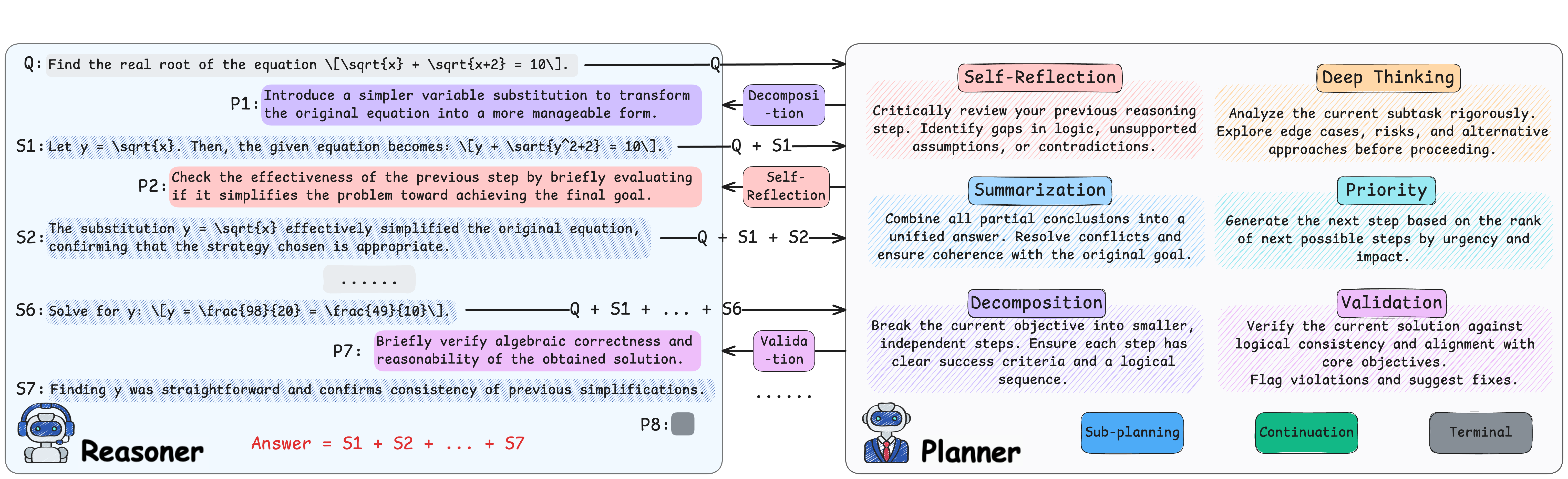} 
    \caption{Illustration of the cooperative pipeline between the \emph{rSIM} planner (leader) and the LLM (reasoner/follower). The planner receives the question and the current reasoning steps, and selects one of nine strategies to inject into the reasoning process to guide the reasoner in generating the next step. This demo is based on a question from the \texttt{MATH} dataset.}
    \label{fig:mainstructure}
    \vspace{-4mm}
\end{figure*}
The planner is designed to guide the reasoner at each reasoning step by selecting one strategy from a predefined set of nine human-designed strategies (Figure~\ref{fig:mainstructure}), which encode core reasoning strategies that LLMs should follow. These strategies, expressed as prompts, help bridge the gap in weak LLMs that inherently lack such capabilities, enabling meaningful improvement during post-training with reinforcement learning. In this framework, the base LLM serves as the reasoner, responsible for generating the next reasoning step based on the planner's selected strategy. This collaboration is naturally modeled as a leader-follower paradigm \cite{leaderfollower-icml23} within a multi-agent RL system. The planner, acting as the leader, takes an action \( \bm{a}^p \sim \pi^p_{\bm{\phi}}\left(\bm{a}^p | \bm{s}\right) \), where \( \pi^p_{\bm{\phi}} \) is the policy parameterized by \( \bm{\phi} \), \( \bm{a}^p \) is a strategy sampled from the action space \( \mathcal{A}^p \), and \( \bm{s}^p \in \mathcal{S}^p\) includes the question and previous reasoning steps. The reasoner, as the follower, then takes an action \( a \sim \pi_{\bm{\theta}}\left(a | \bm{s}, \bm{a}^p\right) \), where \( a \) is the next token and \( \bm{s} \) denotes all tokens generated thus far. 
Eventually, we define the planner's reward as \( R^p = R_{acc} + R_{terminal} + R_{penalty} \) and the reasoner's reward as \( R = R_{acc} + R_{format} + R_{follow} \), where \( R_{terminal} = 1 \) if the final plan is the 'Terminal' strategy, and \( -1 \) otherwise; \( R_{penalty} = - \) (ratio of the most frequently selected strategy); \( R_{follow} \) is the ratio of reasoning steps that follow the given plan; and \( R_{acc} \) and \( R_{format} \) are the accuracy and format rewards as defined in GRPO \cite{GRPO-arxiv24}.

As presented in Figure~\ref{fig:mainstructure}, to answer each question, the two agents interact with each other for \( n \) rounds, where \( n \) corresponds to the number of reasoning steps required to produce the final answer, leading to \( \bm{o}^{dpr} = \left[\bm{p}_1,\bm{z}_1, \bm{p}_2,\bm{z}_2, \dots, \bm{p}_n, \bm{z}_n\right] \), where $\bm{p}_n$ is plan selected by the planner for the n-th reasoning step. In the context of reinforcement learning, we define two advantage functions: \( A^{\pi^p_{\bm{\phi}}}(\bm{s}_t, \bm{a}^p_t) \) for the planner and \( A^{\pi_{\bm{\theta}}}(\bm{s}_t, a_t, \bm{a}^p) \) for the reasoner. Here, \( t \) denotes the index of the generated token. For the planner, we assume that all tokens within a single reasoning step share the same advantage, which is set equal to the plan-level reward. Therefore, we have the following objective $\mathcal{J}_{\bm{o}^{dpr}}$:
\begin{equation}
    \label{eq:main}
    \nonumber
    \begin{aligned}
        &\frac{1}{|\bm{o}^{dpr}|} \sum_{t=1}^{|\bm{o}^{dpr}|} \Bigg[ 
        \lambda \cdot \left( \frac{\pi^p_{\bm{\phi}}(\bm{a}_t^p|\bm{s}_t)}{\pi^p_{\bm{\phi}_{\text{old}}}(\bm{a}_t^p|\bm{s}_t)} \right)
        \cdot A^{\pi^p_{\bm{\phi}}}(\bm{s}_t, \bm{a}_t^p) \\
        + &(1 - \lambda) \cdot \left( \frac{\pi_{\bm{\theta}}(a_t | \bm{s}_t, \bm{a}_t^p)}{\pi_{\bm{\theta}_{\text{old}}}(a_t | \bm{s}_t, \bm{a}_t^p)} \right)
        \cdot A^{\pi_{\bm{\theta}}}(\bm{s}_t, a_t, \bm{a}_t^p) \Bigg]
    \end{aligned}
\end{equation}
where \(\bm{o}^{dpr} \sim (\pi^p_{\bm{\phi}_{old}}, \pi_{\bm{\theta}_{old}})\), and we set \(A^{\pi^p_{\bm{\phi}}}(\bm{s}_t, \bm{a}_t^p) = A^{\pi^p_{\bm{\phi}}}(\bm{s}_i, \bm{a}_i^p)\) if the \(t\)-th token belongs to the reasoning step \(\bm{z}_i\). This means that the reward of the \(i\)-th plan is assigned to all tokens within the reasoning step generated based on that plan.

We follow the GRPO to define the final objective as \( \mathbb{E}_{\left\{\bm{o}^{dpr}_j\right\}_{j=1}^G \sim (\pi^p_{\bm{\phi}_{old}}, \pi_{\bm{\theta}_{old}})} \left[ \frac{1}{G} \sum_{j=1}^G \mathcal{J}_{\bm{o}^{dpr}_j} \right] \), where \( A^{\pi^p_{\bm{\phi}}}(\bm{s}_t, \bm{a}_t^p) = \frac{R^p_j - \text{mean}\left(\left\{R^p_j\right\}_{j=1}^G\right)}{\text{std}\left(\left\{R^p_j\right\}_{j=1}^G\right)} \). 

\subsection{A Two-Stage Training Scheme}
\label{subsec:twostage}

We observe that simultaneously and equally optimizing the policies of both agents often leads to several issues: (1) conflicting policy updates, where the gradients of the two agents may ``pull'' in opposing directions; (2) credit assignment ambiguity, as it becomes unclear whether success or failure is due to the leader's plan or the follower's execution; and (3) competing exploration versus exploitation, where joint exploration may result in catastrophic miscoordination.

Therefore, to enable effective training, we propose a two-stage scheme in which the first stage prioritizes the policy optimization of the planner, while the second stage shifts focus to optimizing the policy of the reasoner, i.e., the base model. Specifically, we ensure a stable training process by adjusting the weighting parameter \( \lambda \) across the two stages. In the first stage, we set \( \lambda = 0.7 \) to emphasize planner optimization, and in the second stage, we reduce it to \( \lambda = 0.3 \) to prioritize the optimization of the reasoner. To keep the main paper concise, further implementation details are provided in the appendix.

\section{Experiments}
\label{sec:exp}

\begin{figure*}[t]
    \centering

    \includegraphics[width=0.95\textwidth]{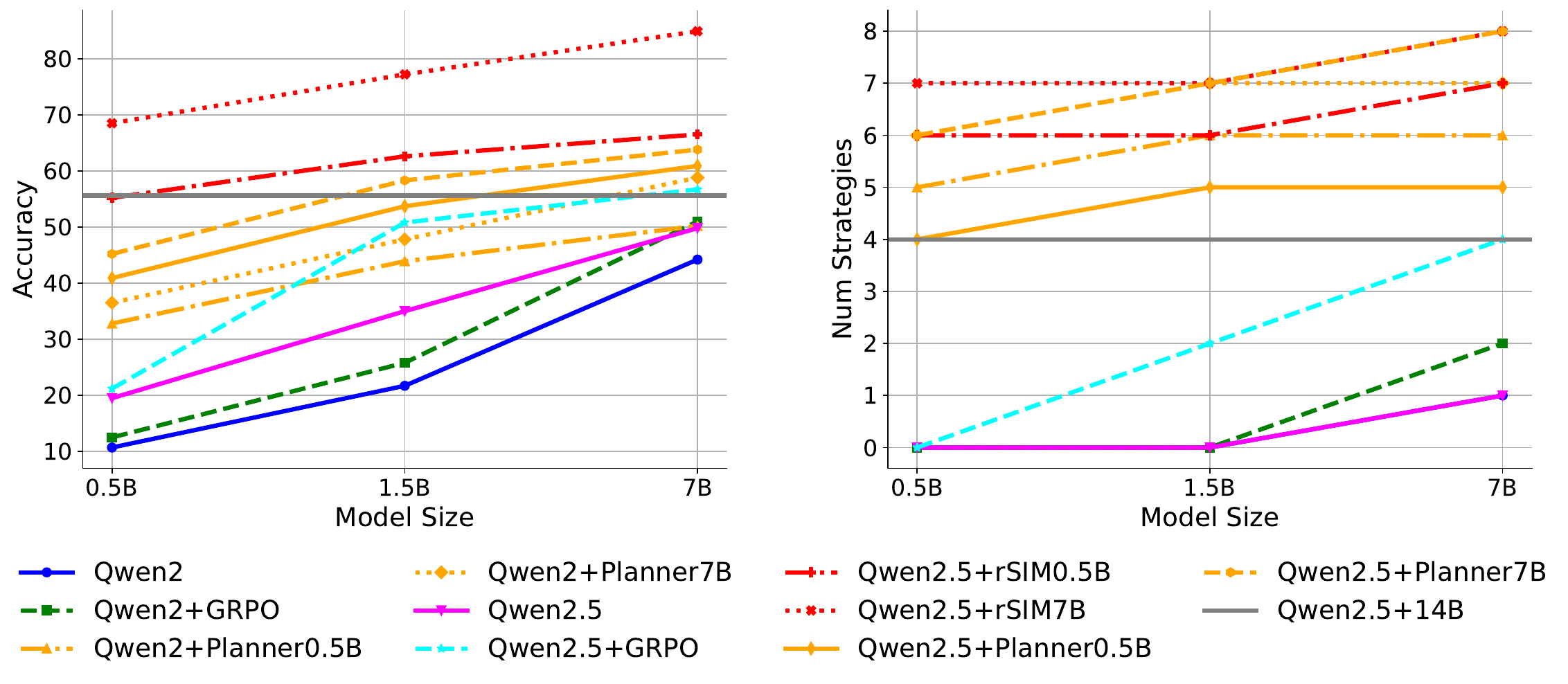} 

    \caption{Accuracy of Qwen2.5 models (0.5B, 1.5B, 7B) on the \texttt{MATH} dataset under various settings. ``+GRPO'' indicates training with GRPO \cite{GRPO-arxiv24}. ``+\emph{rSIM}'' denotes joint training with a Qwen2.5 planner, with the number indicating planner size. ``+Planner'' refers to using trained planners (0.5B, 7B) in plugin mode, derived from the \texttt{MATH} dataset (see Figure~\ref{fig:curves}). Qwen2.5-14B is shown as a baseline with gray horizontal lines.}
    \label{fig:main}
    \vspace{-4mm}
\end{figure*}

\textbf{Datasets}. Experiments are performed on seven datasets from the HuggingFace website: 1). \texttt{MATH} \cite{MATH-arxiv21}, \texttt{GSM8K} \cite{GSM8K-arxiv21}, \texttt{AIME2024} for mathematics, 2). \texttt{MMLU-Pro} \cite{mmlupro-neurips24}, TheoremQA \cite{TheoremQA-emnlp23} for multi-task reasoning, 3) \texttt{CodeAlpaca-20k} \cite{CodeAlpaca20K} and \texttt{HumanEval} for code generation.

\textbf{Training Settings}. Our experiments uses a range of LLMs, including Qwen2 models at 0.5B, 1.5B, and 7B scales, as well as Qwen2.5 models at 0.5B, 1.5B, 7B, and 14B scales. In addition, we incorporate Open-o1 and Deepseek-R1 \cite{DeepseekR1-arxiv25} as base reasoners, which are paired with the trained planner used as a plugin. We use a batch size of 16 with gradient accumulation set to 4. For the GRPO-related hyperparameters, we configure the temperature at 0.9, the maximum prompt and completion lengths at 1024 each, G at 16, and the KL coefficient (beta) at 0.04. The learning rates for the 0.5B, 1.5B, 3B, 7B, and 14B models are 2e-5, 1e-5, 8e-6, 5e-6, and 2e-6, respectively. These settings remain consistent for both the planner and the reasoner. We employ the AdamW optimizer with a cosine learning rate scheduler. During evaluation, we set the planner's temperature to 0 and the reasoner's temperature to 0.3.

\textbf{Baselines}. We compare \emph{rSI} with the recent GRPO method \cite{GRPO-arxiv24}. In addition, we include the Plan-and-Solve (PS+) prompting approach \cite{ps-acl23} and Planner Prompting as two baselines. Specifically, "w/ prompting" in the experiments refers to Planner Prompting, where an LLM is directly prompted to act as the planner and provide the reasoning strategy for each step of generation.

\textbf{Metrics}. The primary evaluation metric is accuracy, defined as the percentage of correct answers. Additionally, to assess the planner's effectiveness as a guidance mechanism, we measure the average number of strategies applied per problem. For the baseline models, we include the candidate strategies defined in our \emph{rSIM} as part of the prompt, allowing the models to select them when appropriate.

\subsection{Main Results}
\label{subsec:dpr}

\begin{table*}[ht]
\centering
\caption{Performance of Llama series models with \emph{rSIM} on four datasets such as TheoremQA, along with comparisons to the Plan-and-Solve (PS+) Prompting method \cite{ps-acl23} and Planner Prompting baseline. When the reasoner (in the Models column) collaborates with a planner, the planner's size is indicated instead of using `No'. When a model name such as Qwen2.5 is specified in the `Planner' column, we perform cross-model evaluation, where Llama serves as the reasoner and Qwen2.5 as the planner. When the "Planner" column specifies `plug-in', the planner is used off-the-shelf without training to guide the reasoner. Moreover, the row containing `w/ prompting' indicates that we directly prompt an LLM to act as the planner. The `w/' specifies the method used to enhance reasoning. The `-' indicates that training did not converge, while the `$\times$' denotes missing results due to untrainable models.}
\label{tab:performance}
\begin{adjustbox}{width=\textwidth, totalheight=\textheight, keepaspectratio}
\begin{tabular}{c|c|c|c|c|c|c|c|c|c}
\toprule
\textbf{Llama3.2} & \textbf{Planner} & \multicolumn{2}{c|}{\texttt{MATH}} & \multicolumn{2}{c|}{\texttt{GSM8K}} & \multicolumn{2}{c|}{\texttt{MMLU-Pro}} & \multicolumn{2}{c}{\texttt{TheoremQA}} \\
\cline{3-10}
\textbf{Models} & & \textbf{Score} & \textbf{\#Strategy} & \textbf{Score} & \textbf{\#Strategy} & \textbf{Score} & \textbf{\#Strategy} & \textbf{Score} & \textbf{\#Strategy} \\
\hline\hline
1B w/ ZeroCoT & No & 30.6 & 0 & 44.4 & 0 & 21.2 & 0 & 13.7 & 0 \\
1B w/ PS+ [4] & No & 28.2 & 0 & 43.7 & 0 & 19.4 & 0 & 12.2 & 0 \\
1B w/ prompting & 3B & 27.4 & 7 & 42.6 & 3 & 16.8 & 8 & 6.6 & 6 \\
1B w/ prompting & 70B & 33.3 & 5 & 46.9 & 3 & 22 & 6 & 14.3 & 5 \\
3B w/ ZeroCoT & No & 48 & 0 & 77.7 & 0 & 30.1 & 0 & 20.7 & 0 \\
3B w/ PS+ [4] & No & 47.5 & 0 & 77.7 & 0 & 30 & 0 & 18.6 & 0 \\
3B w/ prompting & 3B & 46.4 & 7 & 77.1 & 5 & 28.5 & 7 & 19.9 & 8 \\
3B w/ prompting & 70B & 55.5 & 7 & 81.8 & 4 & 31.8 & 5 & 22.8 & 6 \\
1B w/ GRPO & No & - & 0 & - & 0 & $\times$ & 0 & $\times$ & 0 \\
1B w/ \emph{rSIM} & 1B & 57 & 3 & 83.9 & 1 & 30.8 & 4 & 20.9 & 3 \\
1B w/ \emph{rSIM} & 3B & 61.5 & 4 & 86.3 & 3 & 33 & 6 & 25 & 6 \\
1B w/ \emph{rSIM} & Qwen2.5-1.5B & 59.1 & 4 & 84.4 & 1 & 31.8 & 4 & 24.2 & 5 \\\hline
Llama3.3 70B w/ ZeroCoT & No & 77 & 0 & 90.5 & 0 & 68.9 & 0 & 32.3 & 0 \\
Llama3.3 70B w/ PS [4] & No & 78.3 & 0 & 90.9 & 0 & 70 & 0 & 32 & 0 \\
Llama3.3 70B w/ Prompting & 3B & 79.1 & 7 & 90.5 & 4 & 68.9 & 7 & 31.9 & 8 \\
Llama3.3 70B w/ Prompting & 70B & 84 & 6 & 92.9 & 4 & 71.5 & 4 & 38.6 & 5 \\
Llama3.3 70B & 1B plug-in & 83.2 & 3 & 91.7 & 1 & 71.8 & 6 & 39 & 5 \\
Llama3.3 70B & 3B plug-in & 86.3 & 4 & 92.1 & 2 & 72.7 & 5 & 41.8 & 6 \\
Llama3.3 70B & Qwen2.5-1.5B & 83.7 & 4 & 92 & 2 & 72.3 & 6 & 40.7 & 6 \\
\bottomrule
\end{tabular}
\end{adjustbox}
\end{table*}

With \emph{rSIM}, any LLM, especially smaller ones, can be trained to convergence, achieving dramatic improvements in reasoning performance, as shown by the training curve in Figure~\ref{fig:curves} and the high problem-solving accuracy in Figure~\ref{fig:main}. Specifically, for the base Qwen2.5 models in sizes of 0.5B, 1.5B, and 7B, jointly training with a planner, either Qwen2.5-0.5B or Qwen2.5-7B, under the results in significant accuracy gains over the base models, even surpassing stronger base models. For example, all models trained with \emph{rSIM} (tagged as ``+\emph{rSIM}'' in Figure~\ref{fig:main}) outperform their GRPO-trained counterparts (``Qwen2.5+GRPO''). Moreover, the base Qwen2.5-0.5+\emph{rSIM}0.5B model, which employs a 0.5B planner, notably outperforms the Qwen2.5-7B model. Overall, all Qwen2.5+\emph{rSIM}7B models across different base sizes achieve a new state-of-the-art (SOTA) performance. Moreover, as shown in Table~\ref{tab:token_cost}, the planner introduced by our \emph{rSIM} does not significantly increase the token cost when facilitating reasoning in a single LLM.

By comparing the left and right sub-figures of Figure~\ref{fig:main}, we observe that after training under \emph{rSIM}, every base model learns to execute a variety of human-level reasoning strategies as defined by the nine options in Figure~\ref{fig:mainstructure}. In particular, even weaker base models, such as Qwen2.5-0.5B, which initially showed no discernible strategy, can be guided by the planner to employ six or seven strategies during reasoning. These results also reveal a positive correlation between accuracy and the number of strategies used.

More importantly, as shown in Table~\ref{tab:performance}, applying \emph{rSIM} to other types of LLMs, such as a series of Llama models, still yields consistent performance improvements due to the strategy injection provided by the planner. For example, when Llama3.2-1B is used as the reasoner and paired with a Llama3.2-1B planner, the system achieves accuracy close to that of Llama3.3-70B. This suggests that the planner in our DPR framework is transferable across different LLMs, although the degree of improvement may depend on the base model used for planning.

In addition, we present cross-model evaluation results in Table~\ref{tab:performance}, where the trained planner (Qwen2.5-1.5B) is used to assist Llama3.2-1B and Llama3.3-70B by injecting step-wise strategies. The reasoning performance of the Llama models shows an improvement, indicating that the \emph{rSIM} planner possesses cross-model generalization capability.

\subsection{Evaluation of the Pluggable Planner}
\label{subsec:planner}

\begin{figure*}[t]
    \centering
    \begin{subfigure}[b]{0.49\textwidth}
        \includegraphics[width=\linewidth]{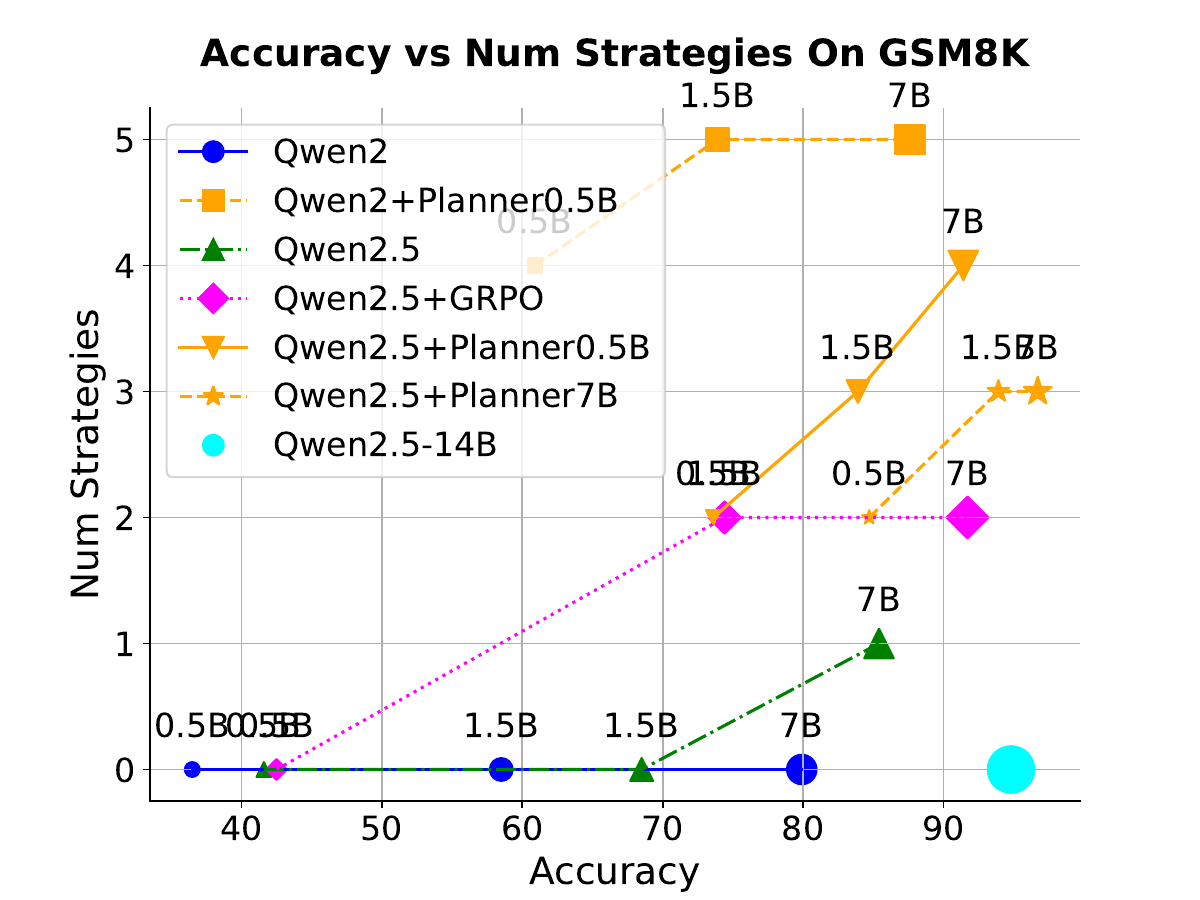} 
        \label{fig:plugin1}
    \end{subfigure}
    \begin{subfigure}[b]{0.49\textwidth}
        \includegraphics[width=\linewidth]{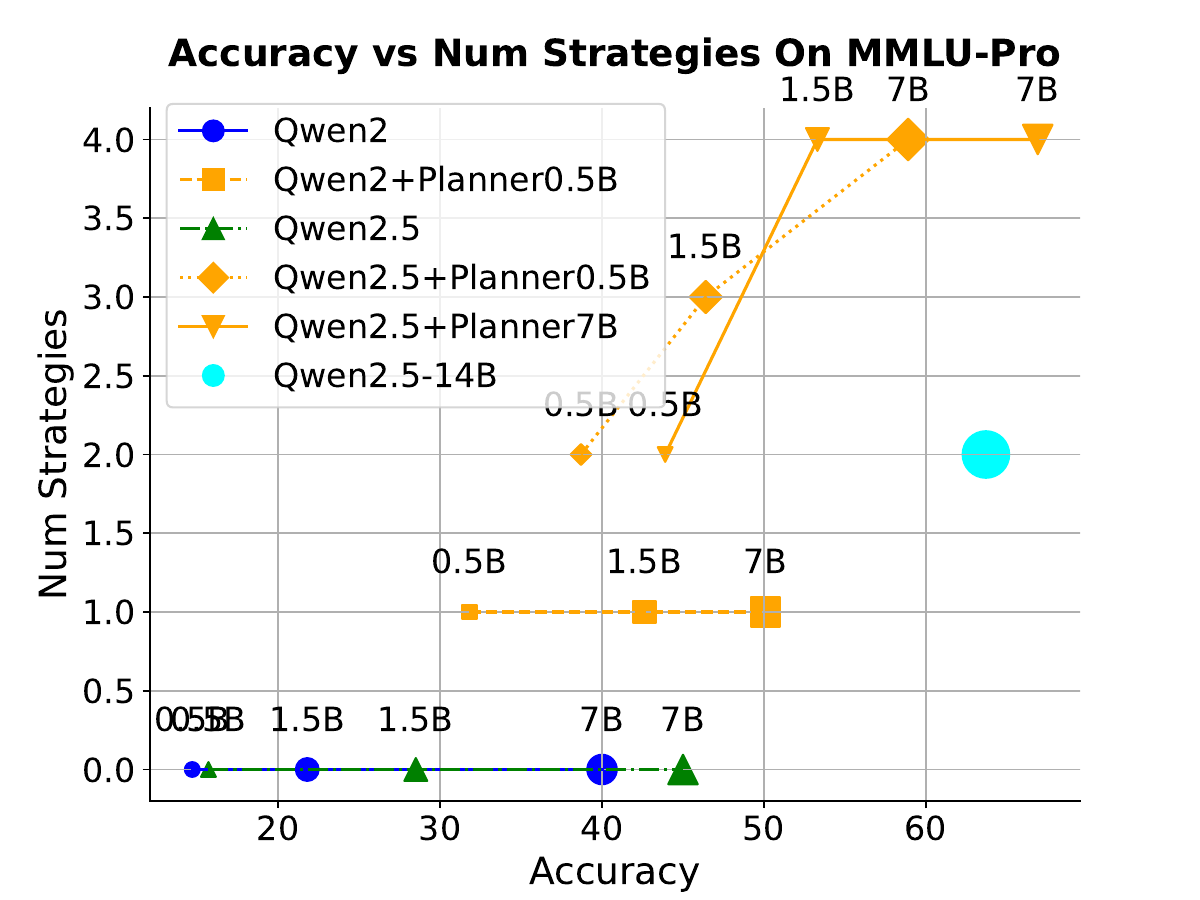} 
        \label{fig:plugin2}
    \end{subfigure}
    \vfill
    \begin{subfigure}[b]{0.49\textwidth}
        \includegraphics[width=\linewidth]{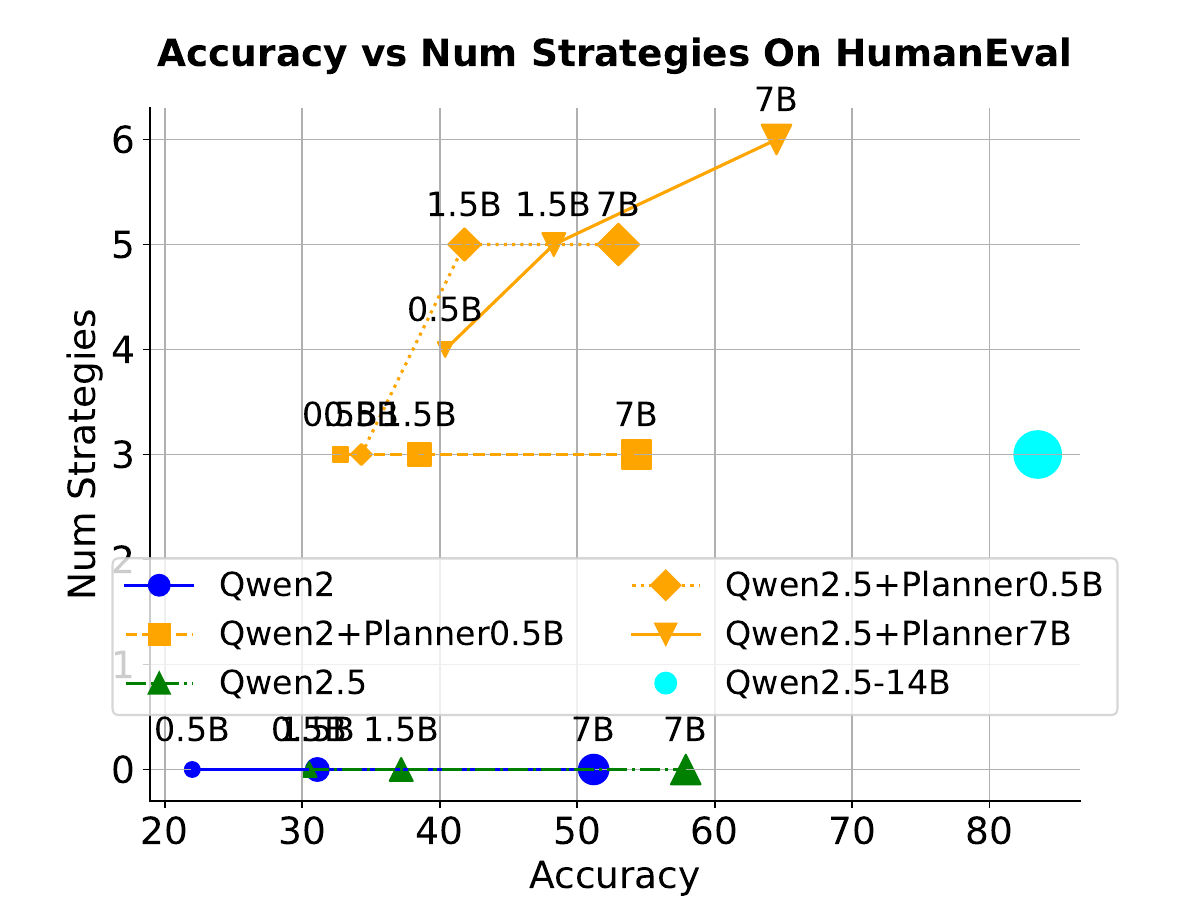} 
        \label{fig:plugin3}
    \end{subfigure}
    \hfill
    \begin{subfigure}[b]{0.49\textwidth}
        \includegraphics[width=\linewidth]{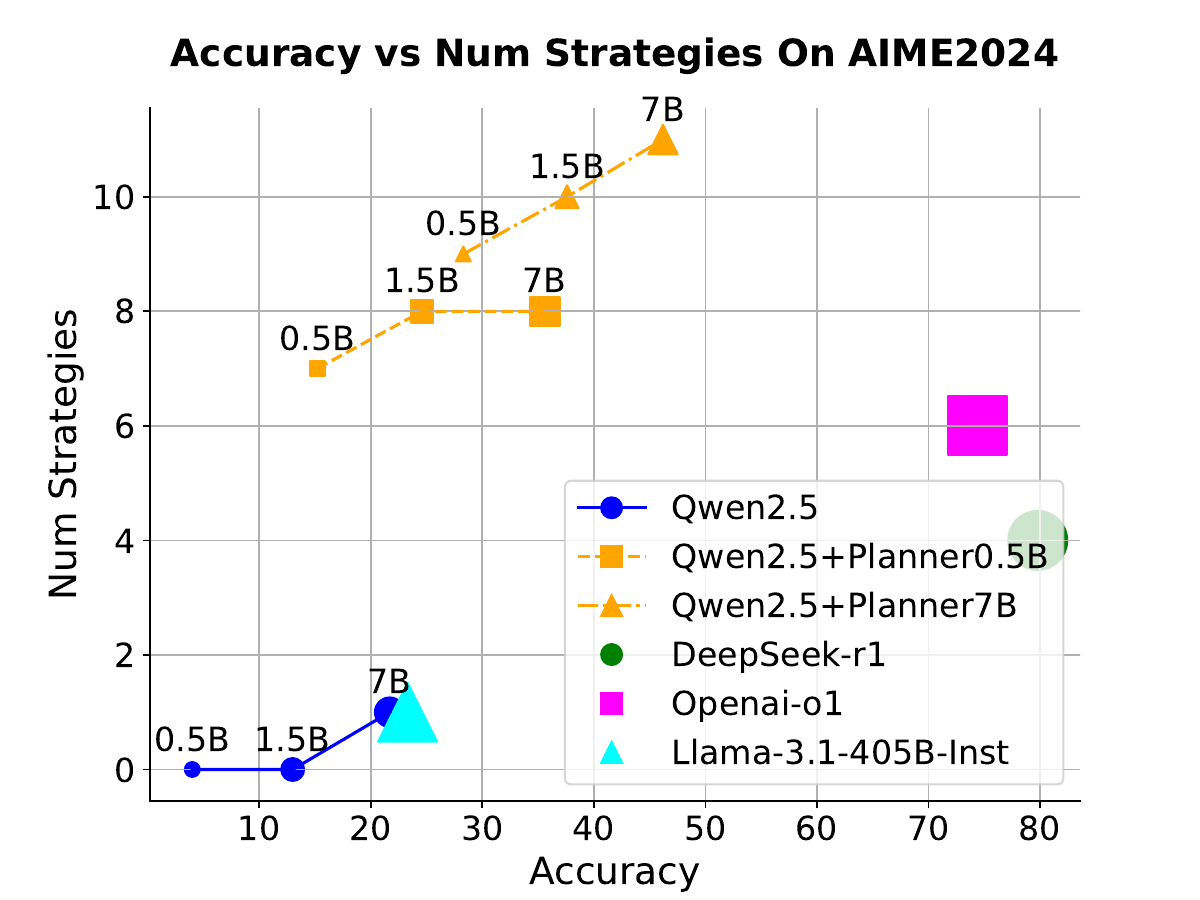} 
        \label{fig:plugin4}
    \end{subfigure}
    \caption{Illustration of accuracy for models of various sizes across different datasets, using the trained \emph{rSIM} planner as a plugin. The terms with ``+Planner'' indicate that the base model collaborates with a trained planner during reasoning. The trained planners are derived from the \texttt{MATH} dataset, as shown in Figure~\ref{fig:curves}.}
    \label{fig:plugin}
    \vspace{-3mm}
\end{figure*}

As shown in Figures~\ref{fig:main} and~\ref{fig:plugin}, which report the performance of base LLMs on five datasets, the trained planner can be used as a plugin to directly introduce human-level reasoning strategies \textbf{without any additional post-training}. In the \texttt{MATH} dataset (Figure~\ref{fig:main}), Qwen2.5+Planner models achieve significantly higher accuracy than both the base Qwen2.5 models and those post-trained with GRPO. A similar trend is observed across all four datasets in Figure~\ref{fig:plugin}: once integrated with the pre-trained planner, base models show substantial accuracy gains and often outperform larger counterparts. For instance, Qwen2.5-1.5B+Planner consistently surpasses the 7B models, and Qwen2.5-7B+Planner achieves SOTA accuracy, even outperforming more powerful models such as Llama-3.1-405B-Inst in the final subfigure of Figure~\ref{fig:main}. More importantly, as shown by the number of strategies used during problem-solving across all datasets, the trained planner enables any LLM to incorporate effective reasoning strategies into the reasoning process. This effect is especially evident in the challenging \texttt{AIME2024} dataset, where the planner enables LLMs to apply human-level strategies more than eight times per problem, resulting in a significant improvement in accuracy. 

These results demonstrate that once planners are jointly trained with even weak reasoners, they can be directly reused to enhance the performance of LLMs on other tasks without any additional post-training. For instance, planners such as Qwen2.5-0.5B and Qwen2.5-7B, trained with the simple Qwen2.5-0.5B reasoner on the \texttt{MATH} dataset, work adaptively with different LLMs across various datasets. This opens a new direction where, instead of fine-tuning LLMs for every task, we can integrate high-intelligence modules such as the planner in \emph{rSIM} to directly boost their reasoning capabilities. In doing so, we shift the focus from optimizing a coupled planning and reasoning process to developing an advanced planner that can collaborate with any LLM.

Similarly, when using Llama series models as base LLMs, the trained planner can serve as a plug-in module to directly inject human-level reasoning strategies. As shown in Table~\ref{tab:performance}, without any additional post-training, Llama3.2-1B, when guided by strategies provided by the Llama3.2-1B or Llama3.2-3B planner, achieves consistent and significant accuracy improvements across all corresponding datasets.

\subsection{Evaluating the Continuous Learning Capability of the Planner}
\label{subsec:plannercl}

\begin{table}
    \centering
    \begin{adjustbox}{width=\columnwidth}
      \begin{tabular}{c|cccccc}
        \toprule
        \textbf{Reasoner} & \multicolumn{1}{c|}{\begin{tabular}[c]{@{}c@{}}\textbf{Planner} \\ (\textit{MATH})\end{tabular}} & \multicolumn{1}{c|}{\texttt{MATH}} & \multicolumn{1}{c|}{\texttt{GSM8K}} & \multicolumn{1}{c|}{\texttt{HumanEval}} & \multicolumn{1}{c|}{\texttt{MMLU\_Pro}} & \texttt{AIME2024} \\ \hline\hline
        \multirow{2}{*}{0.5B} & 0.5B & -0.8 & +0   & +17   & +0.4 & +0 \\
                             & 7B   & +0   & +1.1 & +22.3 & +0.9 & +0 \\ \cline{1-1} \hline
        \multirow{2}{*}{7B}   & 0.5B & +1.2 & +1.8 & +24.4 & +0   & +0 \\
                             & 7B   & +0   & +0.6 & +22.8 & +0   & +0 \\ 
        \bottomrule
      \end{tabular}
    \end{adjustbox}
    
    \caption{Performance gain of the off-the-shelf planner after continued training on the coding task \texttt{CodeAlpaca-20k} under the \emph{rSIM} . With the enhanced planner used as a plugin, the reasoner models (listed in the ``reasoner'' column: Qwen2.5-0.5B and Qwen2.5-7B) address problems across five datasets.}
    \label{tab:continue}
    \vspace{-4mm}
  \end{table}

Table~\ref{tab:continue} shows that the planner can be continuously trained to achieve improved performance on all tasks while preserving its effectiveness on previously seen tasks. Specifically, the off-the-shelf planner, originally optimized on the \texttt{MATH} dataset as shown in Figure~\ref{fig:curves}, is further trained with Qwen2.5-0.5B (serving as the reasoner) using the \texttt{CodeAlpaca-20k} dataset under the \emph{rSIM} to obtain a coding-enhanced planner. This enhanced planner is then used as a plugin with the reasoner models (0.5B and 7B of Qwen2.5) to address questions from five diverse datasets. Table~\ref{tab:continue} reports the performance gains relative to the off-the-shelf \texttt{MATH} planner. Overall, aside from a small drop of 0.8 in accuracy on \texttt{MATH}, the performance of the enhanced planner is either maintained or improved across all tasks. In particular, the improvement on \texttt{HumanEval} is substantial, with gains ranging from 17\% to 24.4\%. This demonstrates that continued training allows the planner to better guide coding-related reasoning using human-level strategies. Such a significant advantage highlights the practical value of the planner, as it can be continuously trained with a small reasoner on mixed-task datasets to improve its \textbf{capabilities in a cumulative way}. It is worth noting that the coding-enhanced planner does not show any accuracy gain on \texttt{AIME2024}.

\section{Ablation Study}
\label{sec:ablation}

In Table~\ref{tab:performance} and Table~\ref{tab:performance_qwen}, we compare the planner of \emph{rSIM} with baseline prompting methods, including PS+ prompting \cite{ps-acl23} and direct planner prompting. From our results, we observe that these prompting-based approaches generally require a powerful language model, such as Qwen2.5 14B, to yield even limited improvements. When the model is smaller, such as LLaMA3.2 3B or Qwen2.5 0.5B, accuracy often decreases, demonstrating the limited reliability and scalability of prompting for plan generation. In addition, as shown in Table~\ref{tab:strategy_ablation} and Figure~\ref{fig:distribution}, we examine the impact of different strategies on reasoning performance across various datasets. The results show that self-reflection is consistently important, while other strategies' effectiveness varies across tasks.

\section{Concluding Remarks}

In this paper, we proposed the reinforced strategy injection mechanism (\emph{rSIM}), which enables any large language model (LLM), including small ones like Qwen2.5-0.5B, to become an advanced reasoning language model (RLM). \emph{rSIM} trains a planner with a reasoner jointly via multi-agent reinforcement learning using a leader-follower algorithm. The planner learns to select the best strategy from a set of nine human-designed options to guide the reasoner step by step. Experiments showed that \emph{rSIM} significantly improves reasoning accuracy across multiple tasks. Importantly, the planner can be used as a plugin without additional training, enabling any LLM to gain advanced reasoning capabilities immediately. The planner also supports continual learning, enabling its reasoning guidance to improve continuously across diverse tasks.

\bibliography{main}

\clearpage
\newpage

\appendix

\section{Implementation Details}
\label{sec:algo}

This section presents the design, training, and evaluation details of our proposed \emph{rSIM}.

\subsection{Finetuning the Reasoner LLM toward Generating Step-wise Reasoning}

We first ensure that the reasoner LLM generates the reasoning process in a step-by-step format. To accomplish this, we adopt the mechanism proposed by \cite{stepverify-iclr23}, fine-tuning the model to use `\textbackslash n\textbackslash n' to clearly separate each reasoning step in the problem-solving process. Specifically, we construct a prompt consisting of five chain-of-thought examples formatted such that each step is separated by `\textbackslash n \textbackslash n'. Using this prompt, we let the reasoner generate reasoning processes randomly on the \texttt{MATH} dataset. To avoid introducing dataset-specific solution information into the model, we select only those generated samples that conform to the desired format but contain incorrect solutions. Ultimately, we obtain 1,000 such samples and fine-tune the reasoner on these samples for one epoch. This process ensures that the reasoning generated by the model consistently follows the intended step-by-step formatting.

\subsection{Counting the Number of Strategies}
\label{subsec:counting}

To determine how many strategies are used by the LLM during reasoning—as shown in Fig. 1(B), Fig. 4, and Fig. 5—we adopt a keyword-matching approach based on strategy names and their synonyms. Specifically, we construct a list of candidate keywords for each of the seven main strategies: \textit{self-reflection}, \textit{decomposition}, \textit{deep thinking}, \textit{validation}, \textit{summarization}, \textit{prioritization}, and \textit{sub-planning}, as follows:

\begin{itemize}
    \item Self-Reflection: review, revisit, reflect, reevaluate, rethink, reexamine, reassess, reconsider, analyze, assess, validate, critique, inspect, examine, audit, diagnose, cross-check.
    \item Decomposition: decompose, break down, divide, split, separate, segment, partition, dissect, analyze, unfold, unwrap, reduce, map out, organize, structure.
    \item Deep Thinking: contemplate, deliberate, reflect, ponder, mull over, reason, deduce, infer, evaluate, scrutinize, meditate, analyze, consider, investigate, explore.
    \item Validation: validate, verify, confirm, check, test, justify, prove, cross-check, ensure, affirm, support, substantiate, corroborate, authenticate, evaluate.
    \item Summarization: summarize, recap, restate, paraphrase, condense, outline, highlight, abstract, generalize, simplify, extract, distill, encapsulate, conclude, report.
    \item Prioritization: prioritize, rank, order, select, choose, emphasize, highlight, focus on, weigh, assess, sort, filter, arrange, allocate, favor.
    \item Sub-planning: plan, outline, design, strategize, organize, arrange, map out, formulate, structure, prepare, coordinate, blueprint, set up.
\end{itemize}

Therefore, once a reasoning step is completed, indicated by the generation of `\textbackslash n\textbackslash n', we identify keywords within that sequence. If a match is found, we increment the count for the corresponding strategy by one.

\subsection{Structure of the Planner}

The planner is a simple decoder-only LLM equipped with an action head—a linear layer that outputs a 9-dimensional vector corresponding to the number of available strategies. The planner takes as input the question and the current reasoning steps. The hidden state of the final token in the sequence is fed into the action head to produce a strategy priority vector. The strategy with the highest score is then selected to guide the reasoner's next reasoning step. For example, when using Qwen2.5-0.5 as the planner, the action head is implemented as a fully connected layer with shape \( 896 \times 9 \). In this case, the hidden vector of the final token in the input sequence serves as the input to the action head for strategy selection.

\subsection{Interactive Sampling Mechanism}

Different from GRPO \cite{GRPO-arxiv24}, which generates \(G\) samples for each question by directly forwarding the question through the LLM \(G\) times, our multi-agent framework adopts an interactive sampling mechanism in which the planner agent and the reasoner agent of \emph{rSIM} interact throughout the reasoning process to generate each sample. We follow the notation introduced in Subsections 3.1 and 4.1, and thus present the detailed procedure in Algorithm Table~\ref{algo: sampling}. It is important to note that when the planner collaborates with an LLM for problem solving, the step-by-step reasoning is generated following the same procedure.

\begin{algorithm}[t]
    \KwInput{Question \( q \), the reasoner policy \(\pi_{\bm{\theta}} \) with its hyperparameters, such as the temperature, the planner policy \(\pi^p_{\bm{\phi}}\) with its hyperparameters, and the number of generations \( G \).}
    \KwOutput{Generated \(G\) samples.} 
    \BlankLine

    Sample \(G\) first strategies from the planner \(\left\{a^p_{1,g}\right\}_{g=1}^G \sim \pi^p_{\bm{\phi}}\left(\cdot|q\right)\) \\

    \textbf{Begin parallel sampling}:

    \For{any \(a^p_{1} \in \left\{a^p_{1,g}\right\}_{g=1}^G\)}{
        \(a^p_{trace} \leftarrow a^p_{1}\), \quad \(\bm{o}^{dpr}=\left[ \right]\), \quad \(\bm{a}^{p}=\left[a^p_{1}\right]\), \quad \(n \leftarrow 1\)  \\
         
        \While{\(a^p_{trace}\) is not \textbf{Terminal}}{
            Autoregressively Decode with \(\bm{a} \sim \pi_{\bm{\theta}}\left(\cdot|q, \bm{o}^{dpr}, a^p_{trace}\right)\) until `\textbackslash n\textbackslash n' is generated\\
            Collect the sequence of decoded \(\bm{a}\) as the reasoning step \(\bm{p}_n,\bm{s}_n \)\\
            Append \(\bm{o}^{dpr} \leftarrow \left[\bm{o}^{dpr}, \bm{p}_n,\bm{s}_n\right]\)\\
            Select strategy action \(a^p \sim \pi^p_{\bm{\phi}}\left(\cdot|q,\bm{o}^{dpr}\right)\)\\
            Set \(a^p_{trace} \leftarrow a^p\), \quad Append \(\bm{a}^{p} \leftarrow \left[\bm{a}^{p}, a^p_{trace}\right]\)\\
            Set \(n \leftarrow n+1\)
            
        }
    }
    Return \(\left\{\bm{o}^{dpr}_{1...G}\right\}, \left\{\bm{a}^{p}_{1...G}\right\}\)
 \caption{Interactive Sampling Mechanism}
\label{algo: sampling}
\end{algorithm}

\begin{algorithm}[t]
    \KwInput{\( q \), reasoner policy \(\pi_{\bm{\theta}} \), planner policy \(\pi^p_{\bm{\phi}}\), dataset \(\mathcal{D}\), $N$.}
    \KwOutput{Optimized \(\bm{\theta}, \bm{\phi}\).} 
    \BlankLine

    \For{\(e=1,\dots,E\)}{
        Set reference models \(\pi_{\bm{\theta}_{ref}} \leftarrow \pi_{\bm{\theta}}\), \quad \(\pi_{\bm{\phi}_{ref}} \leftarrow \pi_{\bm{\phi}}\)\\
        \(lambda \leftarrow 0.7\) \\
        \For{step \(1,\dots,M\)}{
            Sample a batch samples \(\mathcal{D}_b\) from \(\mathcal{D}\)\\
            Update old policy models \(\pi_{\bm{\theta}_{old}} \leftarrow \pi_{\bm{\theta}}\), \quad \(\pi_{\bm{\phi}_{old}} \leftarrow \pi_{\bm{\phi}}\)\\
            Perform \textit{Interactive Sampling Mechanism} to generate \(\left\{\bm{o}^{dpr}_{1...G}\right\}, \left\{\bm{a}^{p}_{1...G}\right\}\) for each question in \(\mathcal{D}_b\)\\ 

            Compute rule-based rewards \(\left\{R^p_j\right\}_{j=1}^G\) and \(\left\{R_j\right\}_{j=1}^G\) for each \(\bm{o}^{dpr}_{j}, \bm{a}^{p}_{j}\) \\
            Compute advantage \(A^{\pi^p_{\bm{\phi}}}(\bm{s}_t, \bm{a}_t^p)\)for the t-th token of each \(j\in G\) through group relative advantage estimation.\\
            
            \(lambda \leftarrow 0.3\) iff. \(step <N\)
            
            Update policy models \(\pi_{\bm{\theta}}, \pi^p_{\bm{\phi}}\) by maximizing the training objective (Equation \ref{eq:main})\\

        }
    }
 \caption{Policy Optimization of Multi-Agent Framework}
\label{algo:traing}
\end{algorithm}

\subsection{Training and Evaluation Details}

Algorithm Table~\ref{algo:traing} presents how to jointly optimize the policies of the two agents. During the \textit{Interactive Sampling Mechanism}, the temperatures for the planner and reasoner are set to be 0.9. We set the maximum prompt length to 2,048 tokens and the maximum generation length to 1,024 tokens for all models. \textbf{It is worth noting that in the training objective (Equation 2), we include only \( \left\{\bm{o}^{dpr}_{1...G}\right\} \) for simplicity, although the full formulation should involve both \( \left\{\bm{o}^{dpr}_{1...G}\right\}, \left\{\bm{a}^{p}_{1...G}\right\} \)}. \textbf{We should note that for tokens generated guided by the action of the planner, this action reward is assigned to each token's reward.}

\textbf{Evaluation setup}. We set the maximum prompt length to 2,048 tokens and the maximum generation length to 1,024 tokens for all models. During evaluation, we adopt the zero-shot setting with a temperature of 0, and report pass@1 (accuracy) using stochastic decoding. When using the planner as a plugin for evaluation, we strictly follow the \textit{Interactive Sampling Mechanism} described in Algorithm Table~\ref{algo: sampling}.

\section{Additional Results}
\label{sec:appendixresults}

\begin{figure}[t]
    \centering
    \begin{subfigure}[b]{0.494\columnwidth}
        \includegraphics[width=\linewidth]{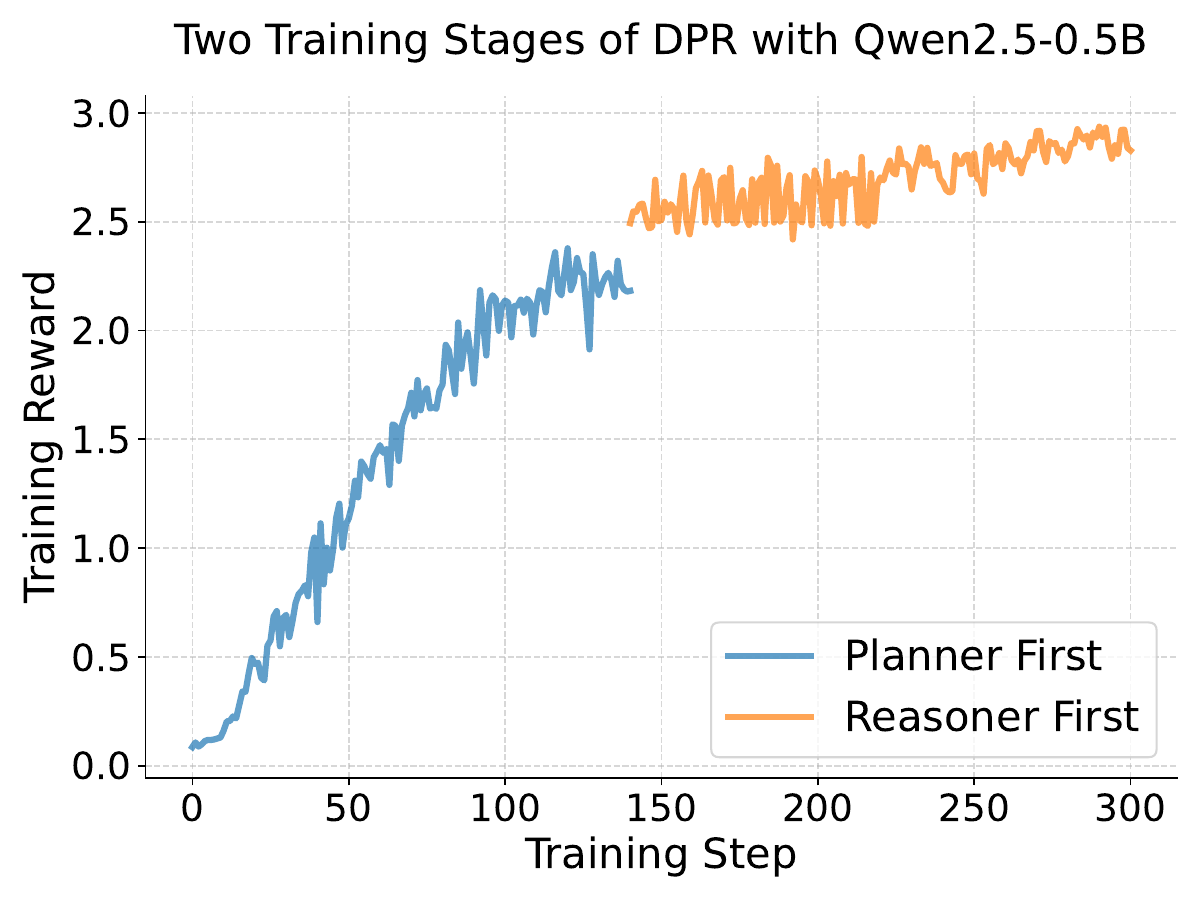} 
        \label{fig:curvetr}
    \end{subfigure}
    \begin{subfigure}[b]{0.494\columnwidth}
        \includegraphics[width=\linewidth]{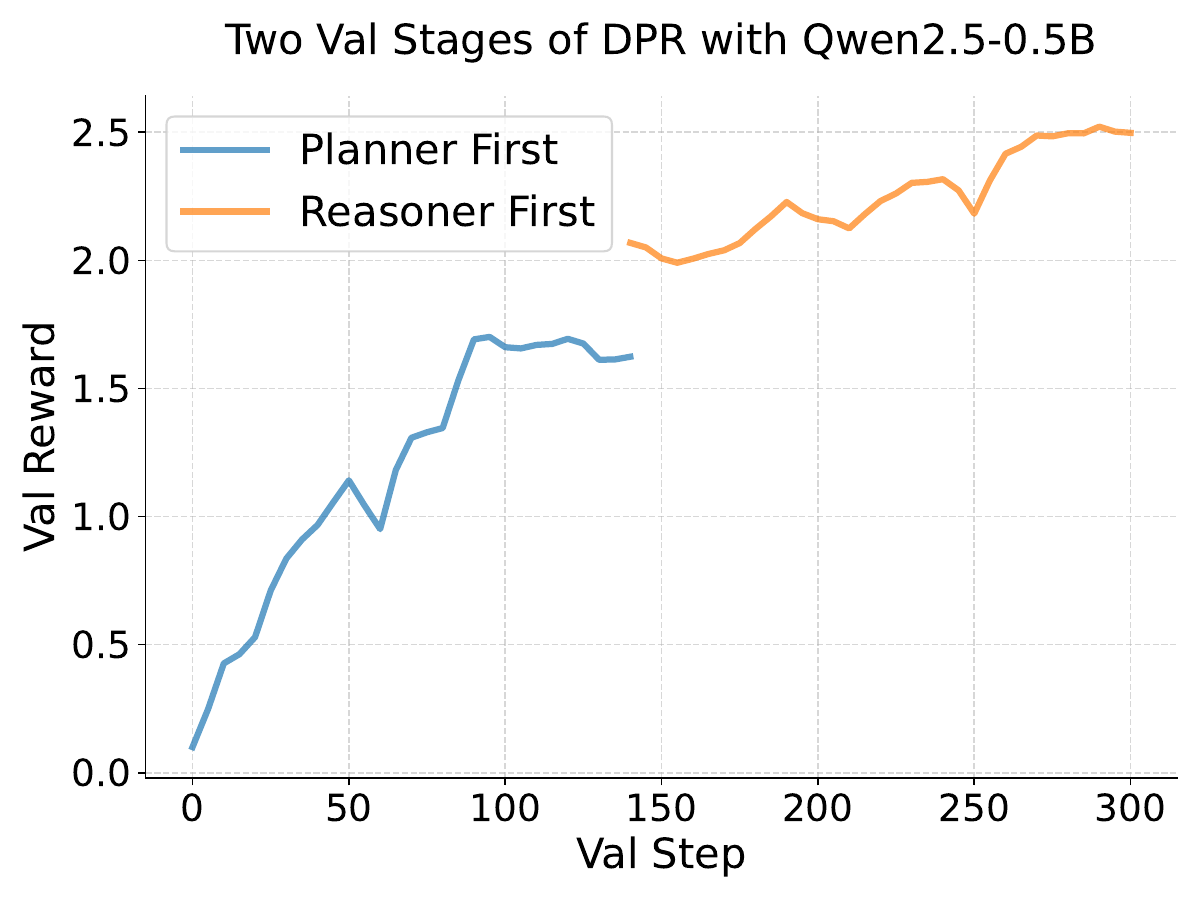} 
        \label{fig:curvets}
    \end{subfigure}
    \vfill
    \begin{subfigure}[b]{0.494\columnwidth}
        \includegraphics[width=\linewidth]{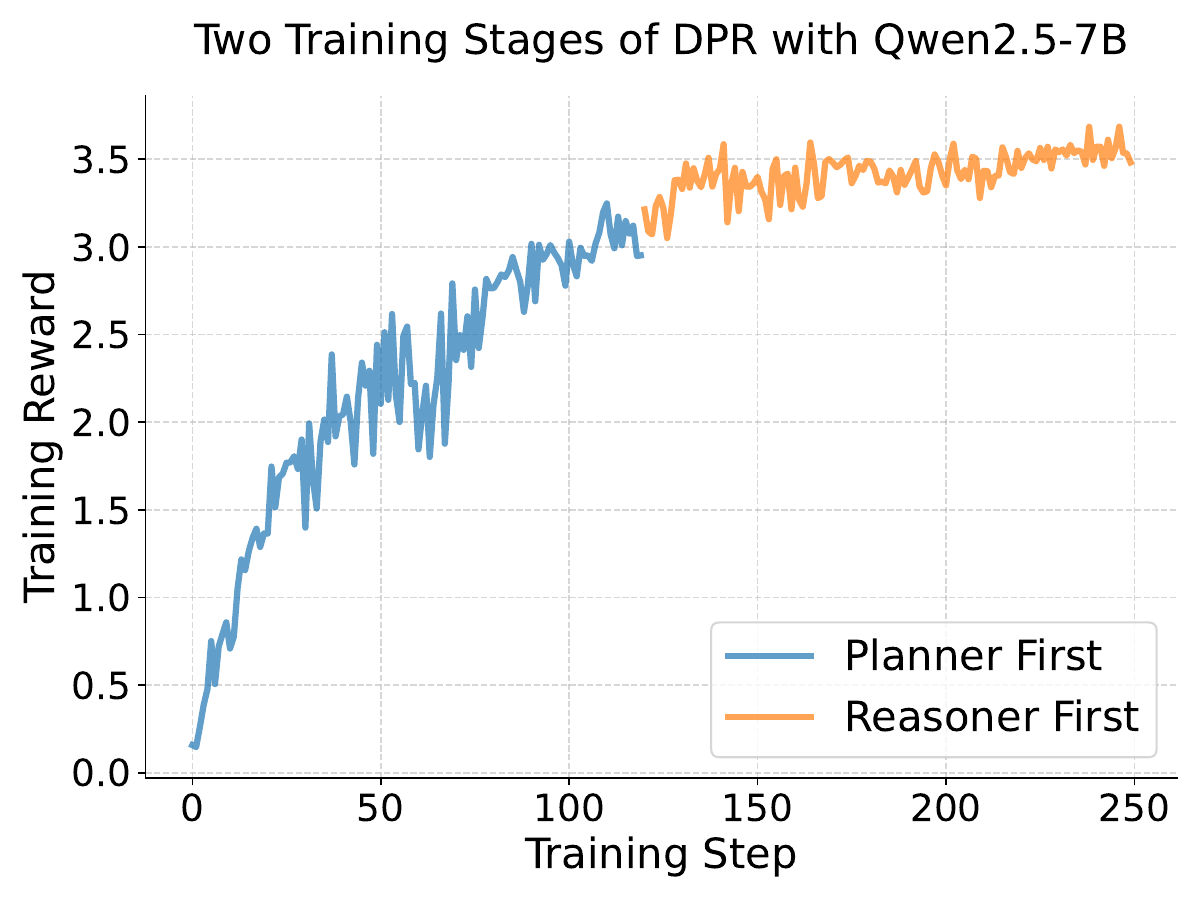} 
        \label{fig:7Br}
    \end{subfigure}
    \hfill
    \begin{subfigure}[b]{0.494\columnwidth}
        \includegraphics[width=\linewidth]{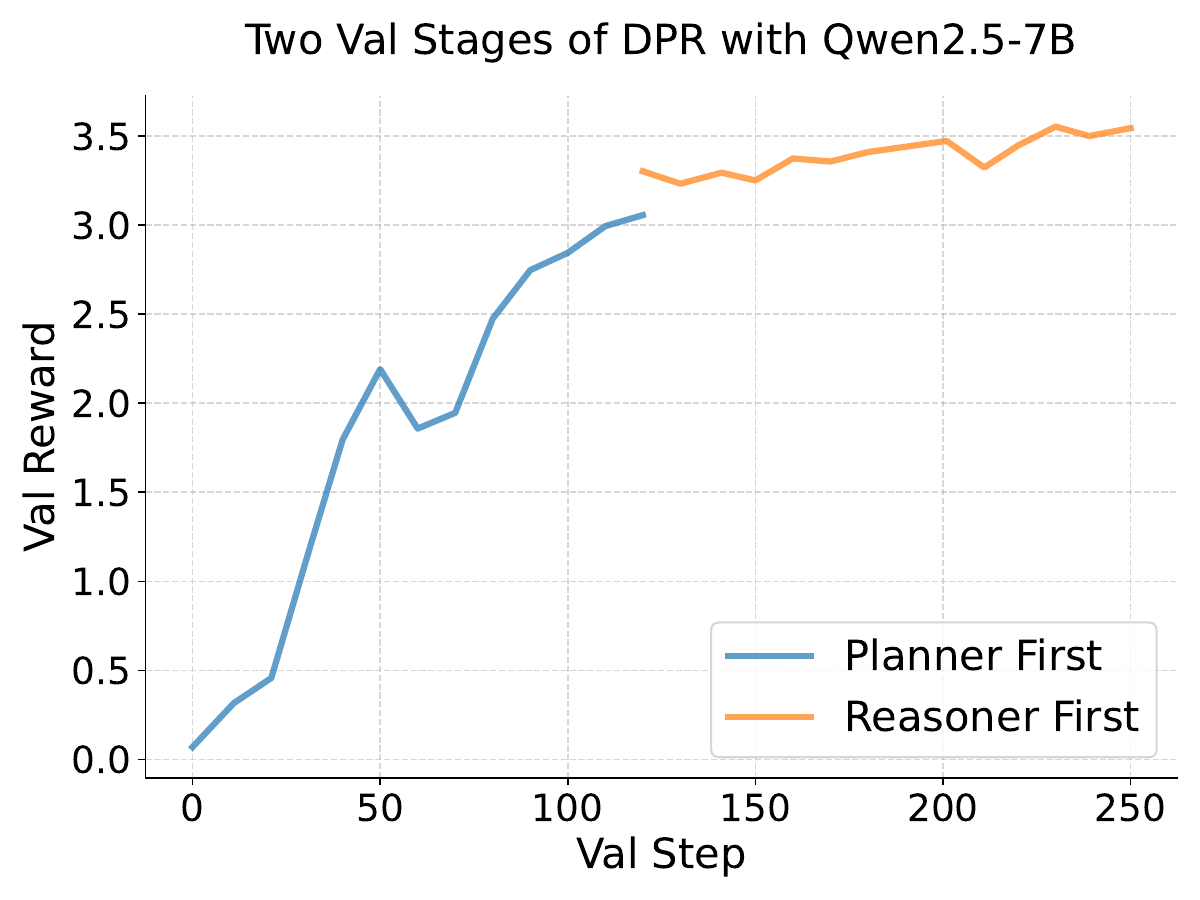} 
        \label{fig:7Bs}
    \end{subfigure}
    \caption{Training and evaluation (Eval) curves of the \emph{rSIM} on the \texttt{MATH} dataset, using either the Qwen2.5-0.5B or Qwen2.5-7B model as the planner, paired with the Qwen2.5-0.5B model as the reasoner. }
    \label{fig:curves}
    \vspace{-4mm}
\end{figure}

As shown in Fig.~\ref{fig:curves}, we train the Qwen2.5-0.5B model, initially lacking high-level reasoning capabilities, using \emph{rSIM} with planners implemented by both Qwen2.5-0.5B and Qwen2.5-7B models. With the two-stage training scheme, total rewards steadily increase until convergence, stabilizing around 2.8 for the 0.5B planner and 3.5 for the 7B planner. Crucially, the planner and reasoner policy models effectively learn to generate and follow step-wise strategies, respectively, resulting in improved reasoning performance. Additionally, the 7B planner achieves higher rewards more rapidly than the 0.5B planner. Consistent training and evaluation trends confirm effective policy optimization. Therefore, we conclude that \textbf{by decoupling planning from reasoning, we can introduce human priors in the form of integrated reasoning strategies into any LLM that lacks them, significantly enhancing reasoning intelligence through our multi-agent RL framework}.

\begin{table}[t]
\centering
\caption{Evaluation of the average generation token cost across different methods on \texttt{MATH} and \texttt{TheoremQA}. We report both the average and standard deviation (mean $\pm$ std) of the total tokens used per question, including tokens used for prompting the LLMs and those generated by the models. This evaluation is conducted on two challenging datasets: \texttt{MATH} and \texttt{TheoremQA}.}
\label{tab:token_cost}
\begin{adjustbox}{width=\columnwidth}
\begin{tabular}{c|c|c|c}
\toprule
\textbf{Methods} & \textbf{Planner} & \texttt{MATH} & \texttt{TheoremQA} \\
\hline\hline
0.5B w/ ZeroCoT & No & 221.6 $\pm$ 172.5 & 250.3 $\pm$ 110.2 \\
14B w/ ZeroCoT & No & 261.8 $\pm$ 192.2 & 308.7 $\pm$ 137.5 \\
0.5B w/ PS+ [4] & No & 327.5 $\pm$ 176.7 & 367.5 $\pm$ 153.6 \\
0.5B w/ Prompting & 14B & 815.2 $\pm$ 356.7 & 993.4 $\pm$ 390.5 \\
0.5B w/ \emph{rSIM} & 7B & 780.3 $\pm$ 210.9 & 800.7 $\pm$ 230.2 \\
7B w/ ZeroCoT & No & 246.9 $\pm$ 189.5 & 291.2 $\pm$ 160.8 \\
7B w/ PS [4] & No & 357.2 $\pm$ 200.9 & 390.6 $\pm$ 190 \\
7B w/ Prompting & 14B & 934.5 $\pm$ 390.2 & 1103 $\pm$ 487.5 \\
7B w/ \emph{rSIM} & 7B & 900.6 $\pm$ 350.8 & 970.2 $\pm$ 427.1 \\
\bottomrule
\end{tabular}
\end{adjustbox}
\vspace{-4mm}
\end{table}

\subsection{Additional Ablation Study}
\label{subsec:appendixablation}

\begin{table*}[ht]
\centering
\caption{Evaluation of the importance of different strategies on reasoning performance when Qwen2.5-0.5 serves as the reasoner and Qwen2.5-7B acts as the planner during inference. In each row labeled with a strategy name, we remove the corresponding strategy from the planner's option set to evaluate its impact on reasoning performance.}
\label{tab:strategy_ablation}
\begin{adjustbox}{width=\textwidth, totalheight=\textheight, keepaspectratio}
\begin{tabular}{c|c|c|c|c|c|c|c|c}
\hline
\textbf{Dataset/Strategy} & \textbf{full} & \textbf{self-reﬂection} & \textbf{deep thinking} & \textbf{decomposition} & \textbf{summarization} & \textbf{validation} & \textbf{prioritization} & \textbf{sub-planning} \\
\hline
\texttt{MATH} & 45.2 & 36.9 & 41.3 & 42.8 & 43.7 & 44.3 & 44.6 & 44.8 \\
\texttt{HumanEval} & 40.2 & 32.3 & 39 & 35.4 & 38.4 & 39 & 39.6 & 35.4 \\
\texttt{MMLU Pro} & 43.9 & 31.3 & 38.7 & 34.5 & 39.6 & 41.8 & 42.7 & 44.1 \\
\texttt{TheoremQA} & 38.7 & 30 & 35.3 & 32.8 & 38.4 & 38.6 & 37.1 & 36.9 \\
\hline
\end{tabular}
\end{adjustbox}
\end{table*}

Table \ref{tab:strategy_ablation} presents the importance of different strategies in improving the reasoning accuracy of Qwen2.5-7B planner in different datasets. Due to time constraints, we do not re-adjust the strategy set for fine-tuning the LLMs across various tasks, nor do we directly compute statistical metrics for each strategy. Instead, we evaluate the importance of each strategy by iteratively masking out one strategy at a time during the evaluation phase. This allows us to observe how the performance is affected when a specific strategy is not considered by the planner. Specifically, when a strategy is masked out, the planner selects the alternative strategy with the second-highest score.

\begin{table*}[ht]
\centering
\caption{Performance of our \emph{rSIM} on datasets such as TheoremQA, along with comparisons to the Plan-and-Solve (PS) Prompting method \cite{ps-acl23} and Planner Prompting baseline. We use Qwen2.5 models in sizes 0.5B, 7B, and 14B. The format and experimental setting of this table are consistent with those in Table 1.}
\label{tab:performance_qwen}
\begin{adjustbox}{width=\textwidth}
\begin{tabular}{c|c|c|c|c|c|c|c}
\toprule
\textbf{Methods} & \textbf{Planner} & \multicolumn{2}{c|}{\texttt{MATH}} & \multicolumn{2}{c|}{\texttt{MMLU-Pro}} & \multicolumn{2}{c|}{\texttt{TheoremQA}} \\
\cline{3-8}
& & \textbf{Score} & \textbf{\#Strategy} & \textbf{Score} & \textbf{\#Strategy} & \textbf{Score} & \textbf{\#Strategy} \\
\hline\hline
0.5B w/ ZeroCoT & No & 19.5 & 0 & 15.7 & 0 & 9.5 & 0 \\
0.5B w/ PS+ [4] & No & 17.2 & 0 & 13 & 0 & 8 & 0 \\
0.5B w/ Prompting & 7B & 21.6 & 7 & 15.9 & 8 & 9.6 & 8 \\
0.5B w/ Prompting & 14B & 26.3 & 6 & 18.6 & 6 & 10 & 5 \\
14B w/ ZeroCoT & No & 55.6 & 0 & 51.2 & 0 & 43 & 0 \\
14B w/ PS+ [4] & No & 57.1 & 0 & 52.3 & 0 & 43.4 & 0 \\
14B w/ Prompting & 7B & 56.8 & 6 & 53.5 & 6 & 43.8 & 7 \\
14B w/ Prompting & 14B & 60 & 5 & 57 & 5 & 47.4 & 5 \\
0.5B w/ \emph{rSIM} & 0.5B & 40.9 & 4 & 38.7 & 2 & 34.1 & 3 \\
0.5B w/ \emph{rSIM} & 7B & 45.2 & 6 & 43.9 & 2 & 38.7 & 4 \\
\hline
7B w/ ZeroCoT & No & 49.8 & 0 & 40 & 0 & 36 & 0 \\
7B w/ PS [4] & No & 49.6 & 0 & 39.2 & 0 & 34.9 & 0 \\
7B w/ Prompting & 7B & 51 & 6 & 40.2 & 8 & 36.4 & 7 \\
7B w/ Prompting & 14B & 56.8 & 5 & 46.7 & 6 & 41.8 & 5 \\
7B w/ \emph{rSIM} & 0.5B & 60.9 & 5 & 58.9 & 4 & 48.3 & 6 \\
7B w/ \emph{rSIM} & 7B & 63.8 & 8 & 66.9 & 4 & 53 & 5 \\
\bottomrule
\end{tabular}
\end{adjustbox}
\end{table*}

\section{Limitations}
\label{sec:limit}

The limitations of our proposed \emph{rSIM} can be categorized into three main aspects: (1) the human-defined action space of the planner is not continuously optimizable and (2) the planner exhibits an imbalanced preference over strategies.

First, the planner's action space is defined solely based on human understanding of the task. As a result, its effectiveness, diversity, and generalization depend heavily on expert priors and cannot be enhanced by the planner itself during reasoning. Without broader exploration during training—particularly in our multi-agent reinforcement learning setting—the \emph{rSIM}'s performance may significantly degrade when necessary strategies are absent from the planner's action space. Furthermore, since each strategy is encoded as a short prompt describing the guidance, the quality and robustness of these prompts directly affect the reasoner's reasoning generation. Misleading, ambiguous, or overly narrow prompts can harm the reasoning process. Unfortunately, such issues are common in human-crafted prompts, including biases and lack of applicability across tasks. More importantly, we currently lack a mechanism to dynamically update or expand the planner's action space. Although this challenge is shared by many reinforcement learning-based approaches—where any change to the action space invalidates the underlying policy—it is particularly critical in our context. Enhancing the strategy space is necessary to improve the generality of LLM-based reasoning frameworks, which is essential for their practical deployment at scale.

\begin{figure}
    \centering
    \includegraphics[width=\columnwidth]{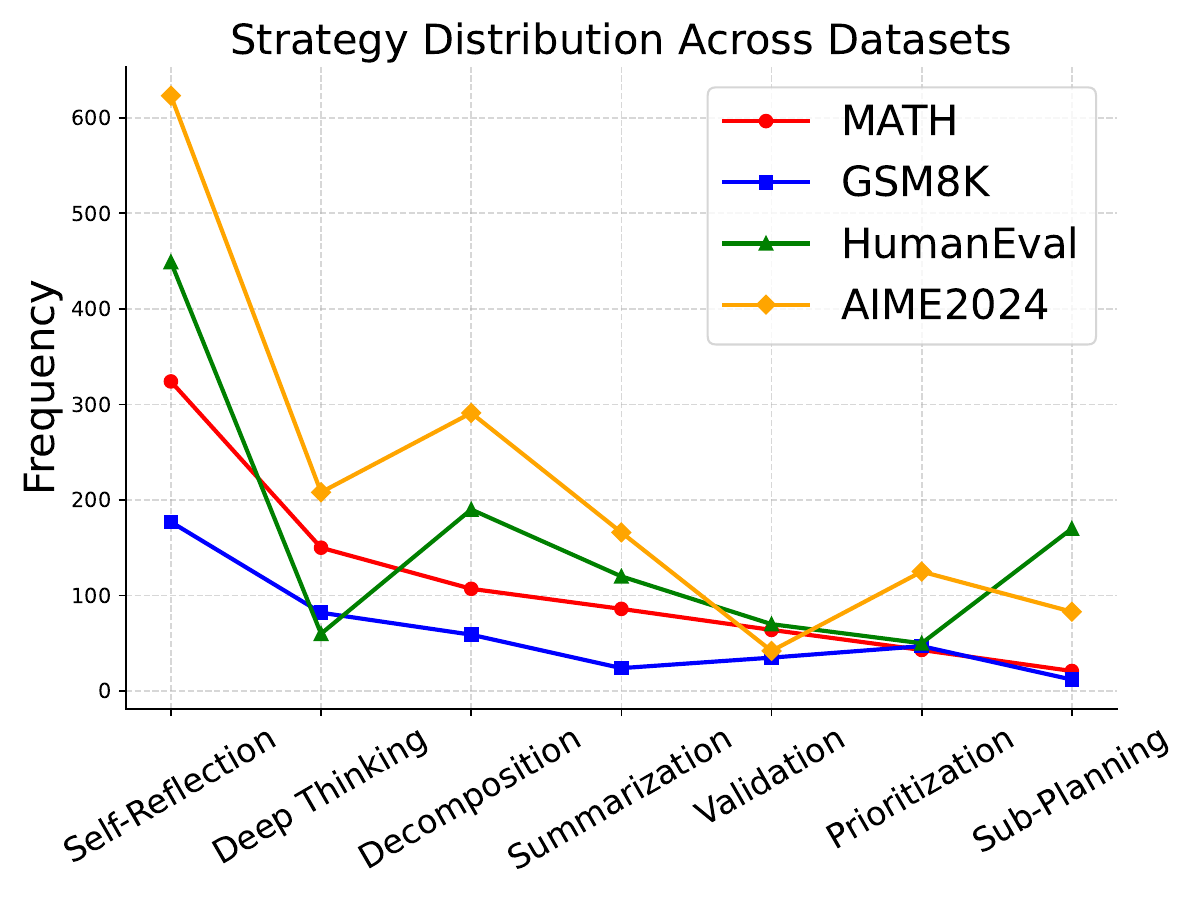}
    \caption{Illustration of how many strategies are used in solving 100 questions from \texttt{MATH}, \texttt{GSM8K}, \texttt{HumanEval}, and \texttt{AIME2024}.}
    \label{fig:distribution}
\end{figure}

Second, we count the total number of strategies used by both the planner (Qwen2.5-0.5B) and the reasoner (Qwen2.5-0.5B) during problem-solving across different datasets. As shown in Fig.~\ref{fig:distribution}, the planner consistently emphasizes the use of \textit{Self-Reflection} across all four datasets, while placing less focus on \textit{Validation}, \textit{Prioritization}, and \textit{Sub-Planning}. Notably, in the \texttt{MATH} dataset, the second most frequently used strategy is \textit{Decomposition}, which is also true for the coding task \texttt{HumanEval}. In addition, for the complex mathematical problems in \texttt{AIME2024}, \textit{Decomposition} again shows a high usage rate, indicating that breaking down the problem during reasoning is crucial for arriving at the correct solution. Therefore, since strategies are not used in a balanced manner during the problem-solving of questions across different tasks, the planner may be unable to comprehensively explore diverse solution paths to reach a reliable answer.

\end{document}